\documentclass{article}

\usepackage[final]{corl_2026}
\makeatletter
\renewcommand{\@conferencelocation}{Austin, Texas, USA}
\makeatother
\usepackage{amsmath,amssymb}
\usepackage{graphicx}
\usepackage{booktabs}
\usepackage{multirow}
\usepackage{enumitem}
\usepackage{wrapfig}

\title{Look Before You Move: Agent-Interface-Aligned Supervision for Open-World Mobile Manipulation}

\author{
  Ronghe QIU \quad Teli MA \quad Jiaming ZHOU \quad Zeying GONG\\
  \textbf{Yiting HE \quad Qiang HUANG \quad Junwei LIANG}\\[4pt]
  The Hong Kong University of Science and Technology (Guangzhou)\\
  {\small\texttt{\{rqiu683,tma184,jzhou760,zgong313,yhe398\}@connect.hkust-gz.edu.cn}}\\
  {\small\texttt{\{qiangh,junweiliang\}@hkust-gz.edu.cn}}
}

\hypersetup{
  pdftitle={Look Before You Move: Agent-Interface-Aligned Supervision for Open-World Mobile Manipulation},
  pdfauthor={Ronghe QIU, Teli MA, Jiaming ZHOU, Zeying GONG, Yiting HE, Qiang HUANG, Junwei LIANG}
}

\begin{document}
\raggedbottom
\maketitle

\begin{abstract}
Open-world mobile manipulation requires a robot to find an object, approach it, and move it to a target receptacle from a language instruction. Training vision-language model (VLM) agents for these tasks requires demonstrations of both action selection and active vision. Existing pipelines often convert planned trajectories into high-level action labels, but shortest-path progress provides limited supervision for deciding where to look. We introduce Agent-Interface Data Alignment (AIDA), a data collection pipeline in which a teacher directly selects and executes actions through the deployed agent's skill interface. For mobile manipulation, we design teachers that maintain target visibility, recover lost targets, and approach for picking or placing. We add template-based spatial reasoning supervision and use a shared harness to organize observations, history, and outputs during training and inference. The resulting agent, AIDA-OWMM, achieves 71.76\% full-task success on 170 held-out OWMM episodes. Teacher sensitivity analysis shows that active-vision demonstrations improve student success. The simulation-trained agent achieves 38\% end-to-end success on a real mobile robot without additional training. Applying AIDA to a navigation subset of EmbodiedBench further demonstrates the pipeline's utility for navigation tasks.
\href{https://connerqiu.github.io/look-before-you-move/}{Project website}.
\end{abstract}


\section{Introduction}
\label{sec:introduction}

Open-world mobile manipulation (OWMM) studies how robots can follow language instructions to find objects and move them to target receptacles in everyday environments~\citep{yenamandra2023homerobot,chen2025owmmagent}. Solving these tasks requires coordinating perception, navigation, and manipulation: the robot must decide where to look, how to approach a target, and when it is ready to pick or place. In an agentic mobile manipulation system, a vision-language model (VLM) selects high-level actions from visual observations, task instructions, and interaction history, while dedicated controllers execute the corresponding motions. A central challenge is training the VLM to make effective decisions as the robot moves, its view changes, and new actions become possible.

Prior work constructs training data by converting planned or executed robot trajectories into high-level action labels~\citep{chen2025owmmagent,internrobotics2025internvlan1}. When these trajectories are driven primarily by shortest-path progress, they can provide limited supervision for active vision: keeping a target visible during approach, recovering it after it leaves view, and obtaining a suitable view for manipulation. This motivates collecting demonstrations from a teacher that actively decides where to look and how to move.

\begin{figure}[t]
  \centering
  \includegraphics[width=\linewidth]{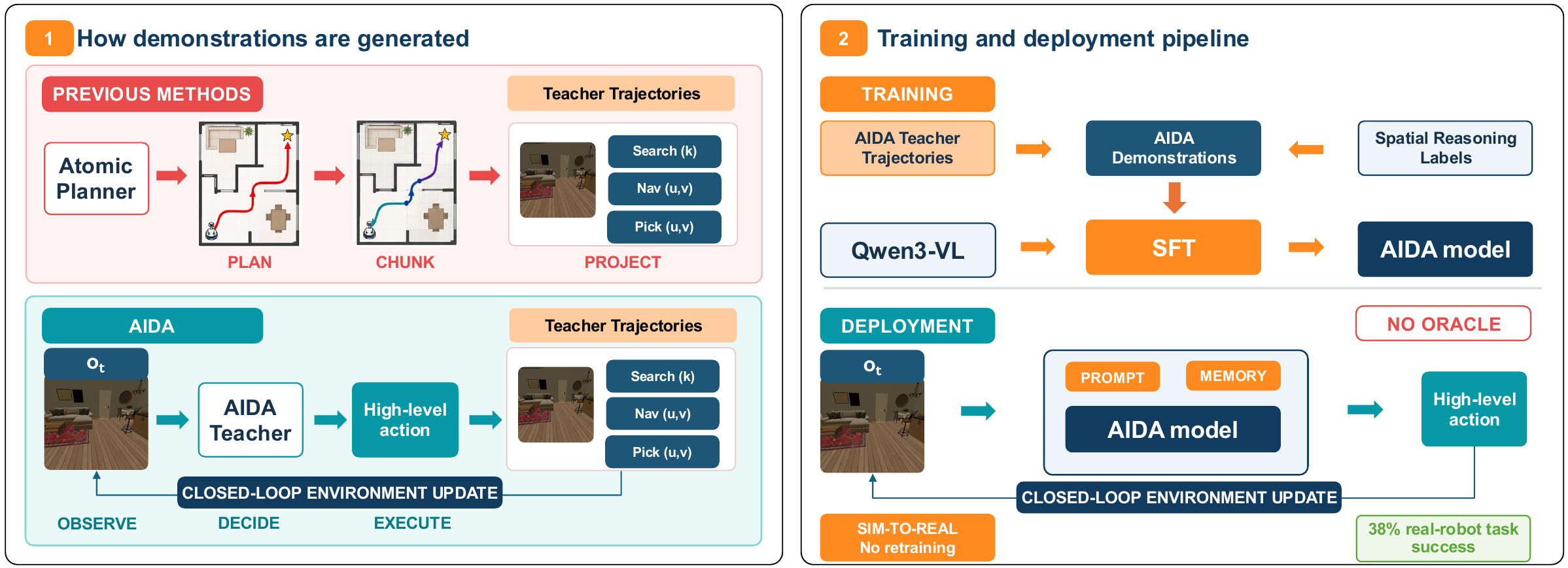}
  \caption{\textbf{AIDA-OWMM overview.}
  Left: the teacher generates demonstrations by selecting and executing high-level actions. Right: spatial annotations and a shared agent harness support training and deployment. The real-robot rate shown includes low-level grasp failures among recorded attempts.}
  \label{fig:motivation}
\end{figure}

To this end, we introduce \textbf{Agent-Interface Data Alignment (AIDA)}, a pipeline for constructing demonstrations through the deployed agent's decision interface, and apply it to mobile manipulation. At each step, the teacher selects a skill and its arguments, executes the call, and uses the resulting observation to choose the next action (Figure~\ref{fig:motivation}). High-level actions thus guide demonstration collection instead of being inferred afterward from motion trajectories. The teacher can use privileged simulator information, while the student's inputs contain only observations and context available at deployment.

Within AIDA, we design the teacher to demonstrate \emph{active target acquisition}. Finding a target is only the beginning: it may leave view during approach or remain out of reach when the robot stops. Building on active perception~\citep{bajcsy2018activeperception,jauhri2023actpermoma}, the teacher moves to collect useful views, turns to recover lost targets, and approaches to prepare for picking or placing. These behaviors teach the student to coordinate looking and moving in preparation for manipulation.

We further provide \textbf{spatial reasoning supervision} to make the judgments behind these actions explicit. For each teacher action, an offline annotator describes whether the target is visible, where it lies relative to the robot, whether the next skill can be executed, and how far the task has progressed. A shared \emph{agent harness} assembles prompts, maintains episode memory, and defines the model's output format consistently during training and deployment. We use supervised fine-tuning to teach the VLM to predict spatial state, reasoning, and actions from its observations and history. At deployment, the model generates these outputs itself.

Our mobile manipulation agent, \textbf{AIDA-OWMM}, achieves 71.76\% full-task success on 170 held-out OWMM episodes. Teacher sensitivity analysis shows that demonstrations from teachers that maintain visibility and recover lost targets yield better student performance than those from a simple approach policy; harness ablations further support action history and spatial reasoning supervision. Trained entirely in simulation, the agent also performs pick-and-place tasks on a real mobile robot in an office environment without additional model training. Finally, we evaluate the approach on short-horizon navigation tasks in EmbodiedBench to examine its applicability beyond the OWMM benchmark.

Our contributions are:\nopagebreak[4]
\begin{itemize}
  \item \textbf{A pipeline for agent-aligned data collection}: AIDA generates teacher demonstrations directly through the deployed agent's high-level action interface.
  \item \textbf{A teacher for active-vision demonstrations}: we design behaviors for maintaining visibility, recovering targets, and approaching for manipulation, and evaluate their effect on student performance.
  \item \textbf{A harness for spatial reasoning supervision}: structured target state and task progress guide action learning through a shared prompting and memory interface.
\end{itemize}

\section{Related Work}
\label{sec:related_work}

\textbf{VLM agents for mobile manipulation.}
Mobile manipulation systems combine semantic task reasoning with physically grounded navigation and manipulation.
Prior work grounds language plans in skill affordances and learns language-conditioned robot policies from large demonstration corpora~\citep{ahn2022saycan,brohan2023rt2,ghosh2024octo,kim2024openvla}, with recent extensions to cross-domain generalization and unseen manipulation tasks~\citep{zhou2025humrobotdomain,ma2025gloverpp,zhou2025agnostos}.
Open-vocabulary mobile-manipulation systems combine language, perception, maps, and robot APIs to complete household tasks~\citep{yenamandra2023homerobot,liu2024okrobot,zhi2025comerobot,lovmm2025,chen2025owmmagent}, while whole-body work highlights stage-dependent perception demands under changing viewpoints~\citep{liu2026incom}.
These systems establish the usefulness of modular high-level skills and open-vocabulary perception.
Our focus is different: we study what form of synthetic supervision teaches the VLM controller to use those skills as a closed-loop active target-acquisition policy.

\textbf{Embodied VLM navigation and view control.}
VLM navigation methods train agents to follow instructions, use visual history, and select navigation actions from partial egocentric observations.
Recent systems study visual history, map use, and action prediction for embodied VLM navigation~\citep{zhang2024navid,zheng2024navillm,zhang2025mapnav,cheng2025navila,wang2023scalevln,wang2025streamvln,han2025roomtour3d}.
Object-goal navigation and active VLN further use spatial abstraction, exploration, and viewpoint correction to search for open-vocabulary targets or reduce route-following uncertainty~\citep{gong2025stairway,wang2020activevln,wei2025dualvln}.
These works show that VLMs can learn route following, memory use, and action prediction, but OWMM target acquisition has a different endpoint: the policy must stop in a robot-object relation from which manipulation can succeed.
AIDA-OWMM therefore trains active view control around task entities and manipulation readiness rather than route progress alone.

Active perception and next-best-view methods choose viewpoints or paths to reveal occluded structure, reduce grasp uncertainty, or improve reachability~\citep{aloimonos1988activevision,bajcsy2018activeperception,morrison2019multiview,breyer2022activegrasp,jauhri2023actpermoma}.
Here, view control is distilled into a VLM high-level policy that decides when to search, turn, approach, pick, and place from the deployed agent interface.

\textbf{Spatial reasoning as policy state.}
Spatial-VLM work studies depth, metric relations, object grounding, cross-view correspondence, and persistent spatial memory~\citep{chen2024spatialvlm,cai2024spatialbot,song2025robospatial,xu2025multispatialmllm,li2025vstp,disjoint3dqa2025,liu2025vlm2}.
Other work turns visual-spatial understanding into points, visual prompts, or multi-turn state-and-transition reasoning for robot actions~\citep{yuan2024robopoint,nasiriany2024pivot,wang2025vagen}.
AIDA-OWMM uses spatial reasoning in a narrower way: object-centric state is an intermediate policy variable for deciding whether to search, turn, approach, pick, or place.
The novelty is using spatial state inside active target-acquisition demonstrations generated in the deployed OWMM agent interface.

\begin{figure}[t]
  \centering
  \includegraphics[width=\linewidth]{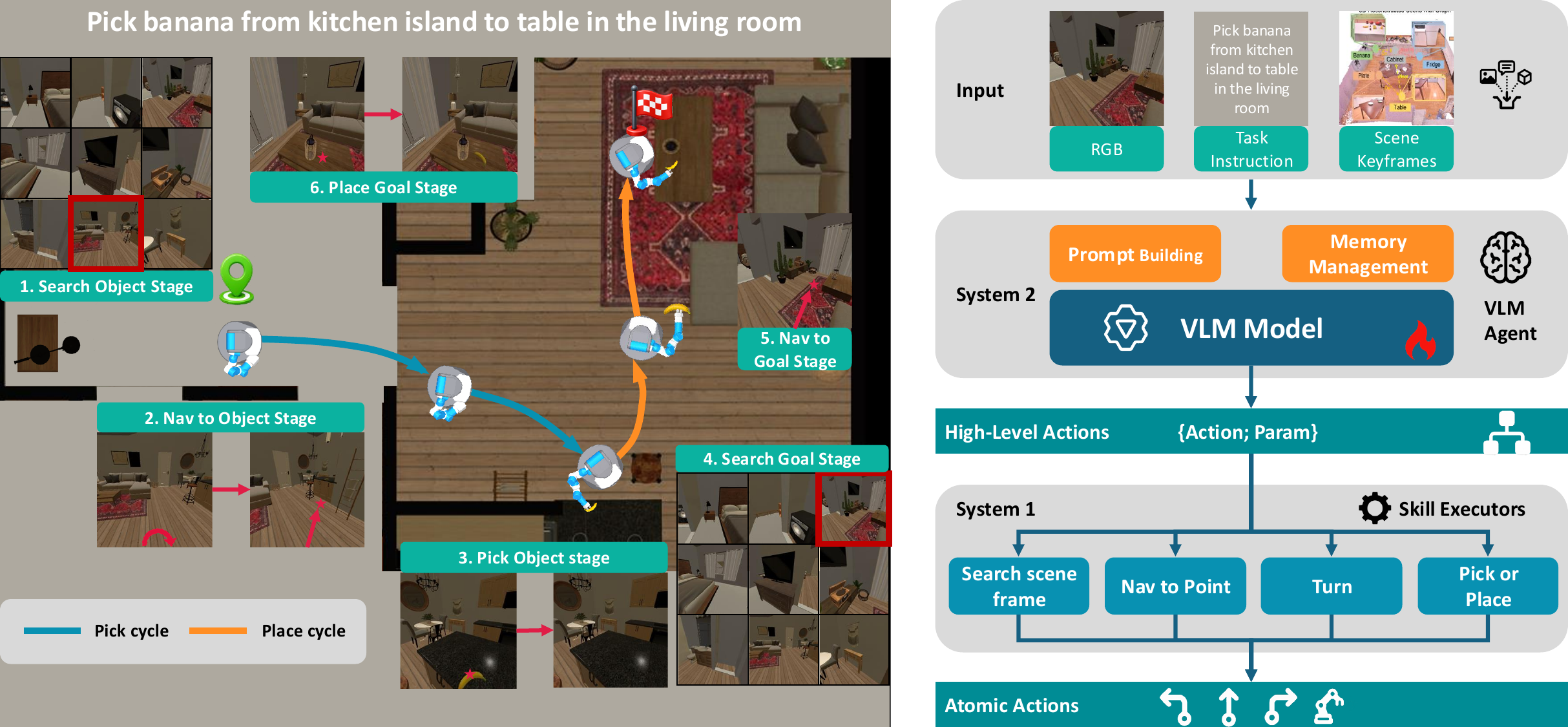}
  \caption{\textbf{OWMM agent interface.}
  The VLM selects search, navigation, turn, pick, and place skills, which fixed controllers execute as robot motions. Scene key frames are provided during search.}
  \label{fig:task}
\end{figure}

\section{Method}
\label{sec:method}

\begin{figure}[t]
  \centering
  \includegraphics[width=\linewidth]{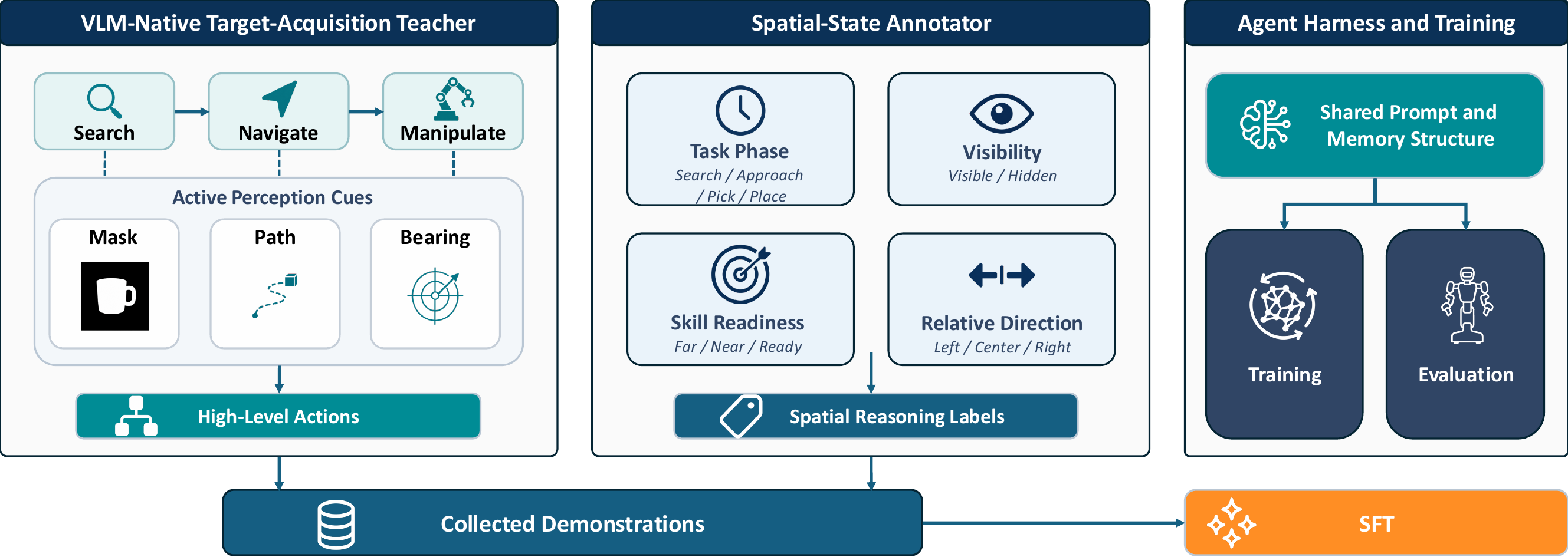}
  \caption{\textbf{AIDA-OWMM pipeline.}
  The teacher collects high-level skill demonstrations, the annotator provides spatial reasoning targets, and the harness constructs training examples and maintains the deployed agent's context.}
  \label{fig:method}
\end{figure}

AIDA-OWMM trains a VLM to select high-level skills for mobile manipulation. The pipeline has three components: a teacher collects demonstrations through the agent's action interface, an annotator adds spatial reasoning targets, and a shared harness organizes prompts and episode memory for training and deployment (Figure~\ref{fig:method}).

\subsection{Problem Formulation}
\label{sec:method:formulation}

An OWMM episode is specified by an instruction $q$, such as moving an object from a source receptacle to a goal receptacle. Following the OWMM benchmark~\citep{chen2025owmmagent}, we assume a pre-mapped environment and known robot localization. The map provides scene geometry for navigation; the agent identifies the object and goal receptacle through RGB key frames with known camera poses and its own visual exploration.

At decision step $t$, the VLM receives $o_t = (I_t^c, q, \mathcal{A}, H_{t-1}, R_t)$, where $I_t^c$ is the current head-camera RGB image, $\mathcal{A}$ is the skill vocabulary, and $H_{t-1}$ contains prior skill calls and their arguments. $R_t$ contains scene key frames during source-object or goal-receptacle search and is omitted during local navigation and manipulation. The VLM outputs $y_t = (r_t, a_t, u_t, m_t)$, where $r_t$ is explicit reasoning about spatial relations and task progress before selecting skill $a_t$ and argument $u_t$, and $m_t$ is a generated step summary that is not included in subsequent history.

Teacher and student share five skills. A fixed executor $\mathcal{E}$ converts each call into low-level robot actions, $\mathbf{a}^{\mathrm{low}}_t = \mathcal{E}(a_t,u_t,I_t^c,D_t^c)$, using the geometric map, robot localization, and stored key-frame poses as context. Here, $D_t^c$ is the depth image used to ground pixel coordinates in 3D. The skills are:
\begin{itemize}
  \item \texttt{search\_scene\_frame} selects a key frame by its displayed ID; the executor uses its associated pose as the navigation destination.
  \item \texttt{turn} specifies an angle for the controller to rotate the robot in place.
  \item \texttt{nav\_to\_point} specifies a normalized pixel coordinate; the executor back-projects it using depth and camera pose to obtain a navigation destination.
  \item \texttt{pick} specifies a normalized pixel coordinate on the object; the manipulation controller moves the end effector toward the corresponding 3D location and grasps the object.
  \item \texttt{place} specifies a normalized pixel coordinate on the destination; the controller moves the held object toward the corresponding 3D location and releases it.
\end{itemize}
In simulation, grasping is simplified: the object is attached to the gripper when the end effector is within the configured grasp distance.

\subsection{Agent-Interface-Aligned Supervision}
\label{sec:method:aida}

AIDA collects demonstrations by executing teacher decisions through the student's skill interface. During collection, the teacher $\pi_T$ uses privileged simulator state $z_t$ to select a skill and its argument, $(a_t^\star,u_t^\star) = \pi_T(I_t^c,q,H^T_{t-1},R_t,z_t,\phi_t)$, where $H^T_{t-1}$ is the observation and action history available during collection and $\phi_t$ denotes the current teacher phase. The selected call is executed, and the resulting observation is used for the next decision. Each recorded action therefore comes from a high-level decision made during the rollout.

\subsection{VLM-Native Target-Acquisition Teacher}
\label{sec:method:teacher}

The teacher uses target visibility, direction, and proximity to choose actions through the VLM's skill interface. It follows the task from source-object search and approach to picking, then goal-receptacle search and approach to placing. Simulator masks, poses, projected paths, and bearings support these decisions during data collection.

During search, the teacher selects the scene key frame whose viewpoint is closest to the target and issues \texttt{search\_scene\_frame}; the search phase completes after this call. During navigation, the teacher actively searches for the target and relevant visual cues, choosing between \texttt{turn} to adjust the view and \texttt{nav\_to\_point} to approach the target. When the target is visible or recently visible, navigation points favor keeping it in view while moving within reach. When the target has not been seen recently, projected paths guide the selection of navigable points. For object pickup during collection, the target must be within 1.0~m with a visible mask exceeding 100 pixels; the teacher uses the $15^\circ$ bearing threshold to align its view. At deployment, the VLM decides when to pick from its observation.

For \texttt{pick} and \texttt{place}, the teacher obtains the object or goal-receptacle mask and uses its centroid, expressed as a normalized pixel coordinate, as the action argument.

A teacher may fail to complete an episode, for example by losing track of the target or exceeding the step budget. Such unsuccessful trajectories are excluded from the training dataset. Teacher thresholds, filtering rules, and recorded fields are detailed in Appendix~\ref{app:method_details}.

\subsection{Spatial-State Annotator}
\label{sec:method:annotator}

After collection, each teacher trajectory provides a sequence of observations and high-level actions. The spatial-state annotator supplements each step with the reasoning that the student is trained to produce before selecting that action. It generates this reasoning offline using deterministic templates and the simulator information recorded during collection.

The templates describe target visibility, relative direction, readiness for the next skill, and task progress, then connect these judgments to the teacher's action. Visibility and direction are derived from the target mask; when the target is outside the image, its direction is obtained from the recorded robot pose and target position. Readiness describes whether the robot should adjust its view, approach, pick, or place, while progress identifies the current part of the task. For example, a template may state that the target is visible on the left but too far away to pick, motivating further approach.

These descriptions form the pre-action reasoning target $r_t$, complementing the recorded skill $a_t$ and argument $u_t$. The resulting supervision teaches the student to reason explicitly about spatial relations and task progress before acting. Label sources and template details are provided in Appendix~\ref{app:method_details}.

\subsection{Shared Agent Harness and Training}
\label{sec:method:harness}

The shared harness uses the same prompt format, history representation, and output schema for training and inference. Each prompt $x_t$ contains the instruction, current RGB image, skill definitions, history $H_{t-1}$, and scene key frames during search. Privileged simulator information is excluded from the prompt in both training and inference.

OWMM-style history recursively carries forward the latest summary~\citep{chen2025owmmagent}. Our history retains previous skill calls and their arguments $(a_i,u_i)$, keeping an explicit record of the agent's actions without including generated step summaries.

The prompt specifies the output format for $y_t=(r_t,a_t,u_t,m_t)$, including the available skills and their argument formats. We fine-tune the VLM on annotated prompt--output pairs using $\mathcal{L}_{\mathrm{SFT}}(\theta) = -\sum_t \log p_\theta(y_t \mid x_t)$. At inference, the harness executes the predicted skill call and updates history and observations for the next decision.

\section{Experiments}
\label{sec:experiments}

Our experiments examine whether AIDA improves mobile manipulation, how teacher behavior and agent supervision contribute to performance, and whether the learned policy transfers to a real robot. We then use a navigation subset of EmbodiedBench to test whether the AIDA pipeline also applies to navigation tasks.

\subsection{Experimental Setup}
\label{sec:exp:setup}

All AIDA policies trained in our experiments use Qwen3-VL-8B-Instruct~\citep{qwen2025qwen3vl} as the base model and are fine-tuned with LoRA using LLaMA-Factory~\citep{zheng2024llamafactory}. The OWMM comparisons vary the demonstration policy, skill set, or harness while keeping the base model, fine-tuning method, and low-level executors fixed.

The teacher sensitivity study uses four skills, excluding \texttt{turn} to match the OWMM action interface. The main AIDA-OWMM model and harness ablations use all five skills: adding \texttt{turn} lets the agent orient toward targets without translating, expanding the viewpoints it can reach. The three AIDA teachers in the sensitivity study each provide 28,480 training examples. The main model and all harness variants use the same larger AIDA dataset of 67,183 examples. Unless varied in the harness ablation, agents use the shared harness described in Section~\ref{sec:method:harness}.

Following the OWMM evaluation protocol~\citep{chen2025owmmagent}, we evaluate on 170 held-out episodes in pre-mapped simulated homes. Table~\ref{tab:owmm_main} reports the fraction of episodes that retrieve a relevant object image (Image Retrieval, Object), pick up the object (Object Picked), retrieve a relevant receptacle image (Image Retrieval, Goal), and approach the receptacle (Robot Close to Goal). Full Task measures successful completion of the object transfer, including placement. Training and evaluation settings are listed in Appendix~\ref{app:owmm_hyperparameters}.

\subsection{OWMM Benchmark}
\label{sec:exp:owmm}
\label{sec:exp:main}

We first compare AIDA-OWMM with existing OWMM systems (Table~\ref{tab:owmm_main}). Results for OWMM-VLM-38B, OWMM-VLM-8B, GPT-4o+PIVOT, and GPT-4o+RoboPoint are taken from OWMM-Agent~\citep{chen2025owmmagent,openai2024gpt4o,nasiriany2024pivot,yuan2024robopoint}. The following studies examine teacher behavior and harness design under the shared training setup.

\begin{table}[h]
  \centering
  \caption{OWMM system comparison. Baseline results are reported by OWMM-Agent; AIDA-OWMM is evaluated on 170 held-out episodes. Intermediate columns report stage completion, and Full Task reports complete object transfer.}
  \label{tab:owmm_main}
  \footnotesize
  \setlength{\tabcolsep}{4pt}
  \resizebox{\textwidth}{!}{%
  \begin{tabular}{@{}llccccc@{}}
    \toprule
    Method & Base model &
    \begin{tabular}[c]{@{}c@{}}Image Retrieval\\(Object)\end{tabular} &
    Object Picked &
    \begin{tabular}[c]{@{}c@{}}Image Retrieval\\(Goal)\end{tabular} &
    \begin{tabular}[c]{@{}c@{}}Robot Close\\to Goal\end{tabular} &
    Full Task \\
    \midrule
    OWMM-VLM-38B & InternVL-38B & 88.56\% & 38.56\% & 30.39\% & 23.53\% & 21.90\% \\
    OWMM-VLM-8B & InternVL-8B & 81.43\% & 17.92\% & 15.96\% & 10.42\% & 9.45\% \\
    GPT-4o + PIVOT & GPT-4o & 59.15\% & 0.65\% & 0.33\% & 0.00\% & 0.33\% \\
    GPT-4o + RoboPoint & GPT-4o & 56.86\% & 1.31\% & 0.00\% & 0.00\% & 0.33\% \\
    \midrule
    \textbf{AIDA-OWMM (ours)} & \textbf{Qwen3-VL-8B-Instruct} & \textbf{90.59\%} & \textbf{82.35\%} & \textbf{74.12\%} & \textbf{72.35\%} & \textbf{71.76\%} \\
    \bottomrule
  \end{tabular}%
  }
\end{table}

AIDA-OWMM achieves 71.76\% full-task success, compared with the reported 21.90\% for OWMM-VLM-38B and 9.45\% for OWMM-VLM-8B. Object-image retrieval is similar for AIDA-OWMM and OWMM-VLM-38B (90.59\% versus 88.56\%), while object pickup reaches 82.35\% versus 38.56\%. The larger difference in pickup shows that the overall improvement extends beyond finding a relevant scene image to approaching and manipulating the object.

\subsection{Agent Sensitivity to Teacher Behavior}
\label{sec:exp:teacher_sensitivity}

AIDA is a data collection pipeline that accommodates teacher policies designed for different embodied tasks. Even within one benchmark, it can collect demonstrations from teachers with different decision rules. Here, we test whether different AIDA teachers improve on trajectory-derived supervision and whether adding active vision to the teacher improves student success.

Table~\ref{tab:teacher_sensitivity} compares four demonstration sources. The OWMM baseline represents trajectory-based data construction: simulator trajectories are divided into chunks, and trajectory points are projected into image coordinates to form high-level action labels. We apply the AIDA annotation and harness to these examples. The other three rows use teachers that directly generate and execute high-level actions through the AIDA pipeline. They share search and pick/place routines but differ in their navigation policy. The \emph{distance-greedy} teacher selects navigable points that reduce distance to the object or receptacle. The \emph{visibility-aware} teacher also selects approach points that keep the target visible after motion. The \emph{target-recovery} teacher includes behavior for finding a previously visible target after it leaves view.

All four students use the same base model, LoRA training method, four-skill interface, spatial annotation, and harness. The OWMM baseline uses 152,478 training examples; each AIDA teacher provides 28,480 examples. We train the three AIDA students with the same seed and report mean success over three evaluations of the 170 episodes.

\begin{table}[t]
  \centering
    \caption{\textbf{Teacher sensitivity.} All students use four skills without \texttt{turn}. Sample counts denote training examples.}
    \label{tab:teacher_sensitivity}
    \small
    \renewcommand{\arraystretch}{0.94}
    \setlength{\tabcolsep}{6pt}
    \begin{tabular*}{0.82\linewidth}{@{\extracolsep{\fill}}lrr@{}}
      \toprule
      Demonstrations & Samples & Success \\
      \midrule
      OWMM, relabeled & 152,478 & 38.8\% \\
      AIDA: distance-greedy & 28,480 & 46.3\% \\
      AIDA: visibility-aware & 28,480 & 65.5\% \\
      AIDA: target recovery & 28,480 & 69.6\% \\
      \bottomrule
    \end{tabular*}
\end{table}

All three AIDA teachers outperform the trajectory-derived baseline: student success rises from 38.8\% for OWMM to 46.3\% for the distance-greedy teacher, 65.5\% for the visibility-aware teacher, and 69.6\% for the target-recovery teacher. These results show that AIDA supports effective supervision from multiple teacher policies. Among the equally sized AIDA datasets, demonstrating active visual behavior further improves task completion.

\subsection{Harness Ablations}
\label{sec:exp:harness}

We next examine how the harness presents the demonstrations to the agent. Table~\ref{tab:harness_summary} holds the AIDA trajectories fixed and varies scene key-frame context, action history, and reasoning supervision. Each row uses its specified format for both training and inference.

\begin{table}[t]
  \centering
\caption{\textbf{Harness ablations.} All rows use the same AIDA demonstrations and five skills. Reasoning denotes spatial reasoning supervision.}
\label{tab:harness_summary}
\small
\setlength{\tabcolsep}{6pt}
\begin{tabular*}{0.82\linewidth}{@{\extracolsep{\fill}}llcr@{}}
  \toprule
  Key frames & History & Reasoning & Success \\
  \midrule
  Every stage & None & No & 55.88\% \\
  Search only & None & No & 60.00\% \\
  Search only & Actions & No & 65.88\% \\
  Search only & Actions & Yes & 71.76\% \\
  \bottomrule
\end{tabular*}
\end{table}

Providing key frames only during search improves success from 55.88\% to 60.00\% in the configuration without history or reasoning. Adding action history raises success to 65.88\%, and adding spatial reasoning supervision reaches 71.76\%. The results favor providing scene key frames during search and retaining past actions to support subsequent decisions.

\subsection{Real-Robot Evaluation}
\label{sec:exp:real}

To test transfer from simulation to physical mobile manipulation, we deploy AIDA-OWMM on an Atribot S1 without further model training. The office environment contains two office rooms and an open tea room, with five movable objects and five receptacles. The agent uses RGB observations and scene key frames through the same high-level skill interface; the robot's controllers execute navigation and manipulation. Hardware and execution details are given in Appendix~\ref{app:experiment_details}.

The robot achieves 38\% end-to-end success with a 15-step horizon, including low-level grasp failures in the evaluation. This result demonstrates transfer from simulation to physical mobile manipulation without additional model training.

\begin{figure}[t]
  \centering
  \includegraphics[width=\linewidth]{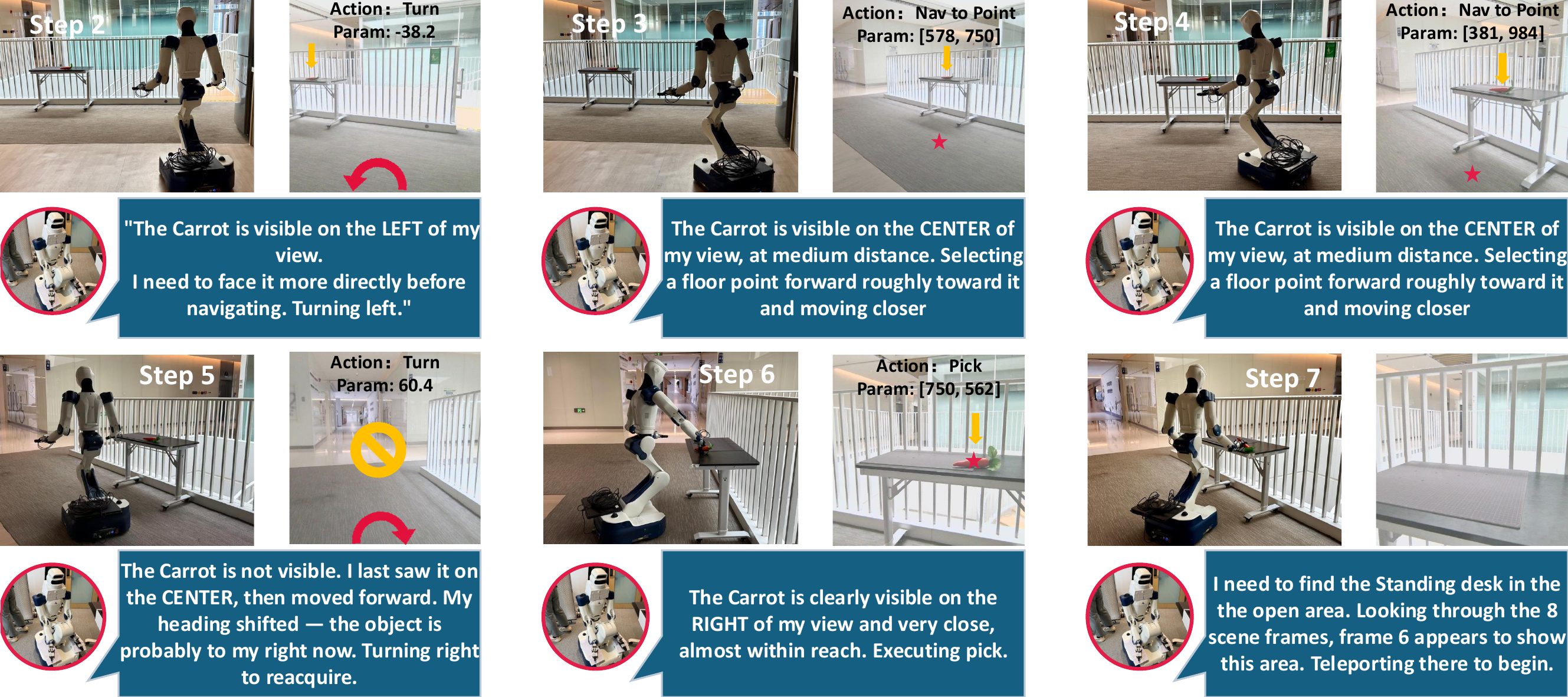}
  \caption{\textbf{Real-robot rollout.}
  The simulation-trained agent turns to locate the object, approaches using pixel-coordinate goals, picks it, and searches for the destination receptacle.}
  \label{fig:real_qualitative}
\end{figure}

Figure~\ref{fig:real_qualitative} shows the resulting sequence of viewpoint changes, navigation, and manipulation. Failure analysis is provided in Appendix~\ref{sec:exp:failures}.

\subsection{EmbodiedBench Navigation}
\label{sec:exp:embodiedbench}

To test whether the AIDA pipeline extends beyond mobile manipulation, we apply it to the navigation subset of EmbodiedBench~\citep{yang2025embodiedbench} used in our evaluation, which consists of language-conditioned, single-room tasks in AI2-THOR. We train a policy on 1,200 navigation episodes constructed with the AIDA pipeline. We also include the OWMM-trained policy as a zero-shot reference. Evaluation uses 60 episodes from each of the \texttt{base} and \texttt{common\_sense} splits, with results averaged over two runs. Hosted-model baselines use the same prompt format, action parser, skill executor, and 10-step horizon. Evaluation protocol details are provided in Appendix~\ref{app:experiment_details}.

\begin{table}[t]
  \centering
\caption{\textbf{EmbodiedBench navigation.} The EB-trained policy tests the AIDA pipeline on navigation; the OWMM-trained policy is a zero-shot reference.}
\label{tab:embodiedbench}
\small
\setlength{\tabcolsep}{6pt}
\begin{tabular*}{0.82\linewidth}{@{\extracolsep{\fill}}lccc@{}}
  \toprule
  Model & Base & Common & Avg. \\
  \midrule
  \multicolumn{4}{@{}l}{\textit{Zero-shot VLMs}} \\
  GPT-5.5 & 0.78 & 0.77 & 0.78 \\
  GPT-4o & 0.43 & 0.37 & 0.40 \\
  OpenAI o3 & 0.53 & 0.50 & 0.52 \\
  Gemini 3 Flash & 0.08 & 0.15 & 0.12 \\
  Claude Sonnet 4.6 & 0.27 & 0.33 & 0.30 \\
  Qwen3-Max & 0.07 & 0.10 & 0.08 \\
  Qwen3-VL-8B-Instruct & 0.02 & 0.02 & 0.02 \\
  \midrule
  \multicolumn{4}{@{}l}{\textit{AIDA policies}} \\
  \textbf{AIDA (EB-trained)} & \textbf{0.85} & \textbf{0.83} & \textbf{0.84} \\
  AIDA (OWMM-trained) & 0.54 & 0.55 & 0.55 \\
  \bottomrule
\end{tabular*}
\end{table}

The EmbodiedBench-trained policy reaches 0.84 average success, while the OWMM-trained policy reaches 0.55 without additional training (Table~\ref{tab:embodiedbench}). The zero-shot Qwen3-VL-8B-Instruct base model reaches 0.02 under the same evaluation interface. The performance of the EmbodiedBench-trained policy supports applying the AIDA pipeline to navigation beyond the mobile manipulation setting.

\section{Conclusion}
\label{sec:conclusion}

We presented AIDA, a pipeline that collects demonstrations through the high-level skill interface used by the deployed agent. For mobile manipulation, we combine active-vision teachers with spatial reasoning supervision and a shared training and inference harness. The pipeline supports different teacher policies, with visibility-aware approach and target recovery improving student success. Real-robot deployment without additional training and the application to a navigation subset of EmbodiedBench extend the evidence beyond simulated mobile manipulation. Together, these results highlight the value of designing demonstrations around the decisions an embodied agent must make, including where to look and how to move before manipulation.

\section{Limitations}
\label{sec:limitations}

AIDA-OWMM uses an offline teacher with privileged simulator state, so generalization is bounded by the teacher data diversity across scenes, objects, embodiments, sensors, and skills.
The OWMM simulator abstracts contact-rich grasping, slip, force control, and recovery; real-robot results show that low-level pick reliability remains a bottleneck.
For navigation, our evaluation is limited to relatively short episodes in the evaluated navigation subset of EmbodiedBench.
Whether agent-interface-aligned supervision provides similar benefits for longer instruction-following and search-based navigation settings, such as VLN-CE~\citep{krantz_vlnce_2020} and ObjectNav~\citep{batra2020objectnav}, remains an important direction for future work.

\clearpage
\acknowledgments{We thank the reviewers and collaborators for helpful feedback.}

\bibliography{references}

\clearpage
\appendix
\section{Method Details}
\label{app:method_details}

\noindent\textbf{Teacher navigation rules.}
The teacher turns when the target or path-segment bearing exceeds $15^\circ$. Heading errors above $40^\circ$ are split across multiple turns. When aligned, the teacher selects a \texttt{nav\_to\_point} goal. For a visible or recently visible target, it selects floor points that preserve visibility and approach a reachable region around the object or receptacle. Otherwise, it samples floor points near the projected short-horizon path, favoring points away from navigability boundaries. 

\noindent\textbf{Trajectory selection and records.}
For the full AIDA dataset, collection attempts cover 13,608 episodes, of which 11,740 succeed. Removing 1,528 successful episodes longer than 16 steps leaves 10,212 episodes and 67,183 training examples. Each step stores the RGB observation, instruction, scene-key-frame order, task phase, skill, and argument. Additional simulator records---robot pose, target positions, mask statistics, turn mode, and navigation pixel---support offline annotation and diagnostics.

\noindent\textbf{Annotation and output format.}
The templates use masks for visible-target descriptions and recorded poses for out-of-view direction labels (Section~\ref{sec:method:annotator}). The five output fields map to the agent representation as follows: \texttt{state\_estimation} and \texttt{reasoning} together represent the pre-action reasoning $r_t$; \texttt{action} and \texttt{action\_information} encode $(a_t,u_t)$; and \texttt{summarization} stores the step summary $m_t$.

\section{Additional Evaluation Details}
\label{app:experiment_details}

\noindent\textbf{EmbodiedBench.}
We use the navigation subset from the VAGEN/EmbodiedBench split~\citep{wang2025vagen,yang2025embodiedbench}. Hosted models share the prompt, output schema, parser, invalid-action handling, executor, and 10-step horizon. Scores average two runs; Table~\ref{tab:embodiedbench} lists public model family names~\citep{openai2026gpt55,openai2024gpt4o,openai2025o3,google2025gemini3flash,anthropic2026claudesonnet46,qwen2025qwen3max,qwen2025qwen3vl}.

\noindent\textbf{Real-robot perception and navigation.}
We map the office with FAST-LIO2~\citep{xu2022fastlio2} and manually select 15 scene key frames. The VLM receives RGB observations from the torso RGB-D camera. The \texttt{search\_scene\_frame} skill navigates to the selected key-frame pose. For \texttt{nav\_to\_point}, the executor back-projects the selected pixel using depth and plans an obstacle-free path with the fast marching method.

\noindent\textbf{Real-robot manipulation.}
The executor expresses each pick or place target in the robot base frame and sends an end-effector pose command to the Atribot whole-body controller. Grasping uses a predefined pose. Near the object, head-camera visual servoing uses SAM~3~\citep{carion2025sam3} to refine the grasp position. Low-level grasp failures count toward the reported end-to-end failure rate.

\section{Failure Analysis}
\label{sec:exp:failures}

\begin{figure}[!htbp]
  \centering
  \includegraphics[width=0.70\linewidth]{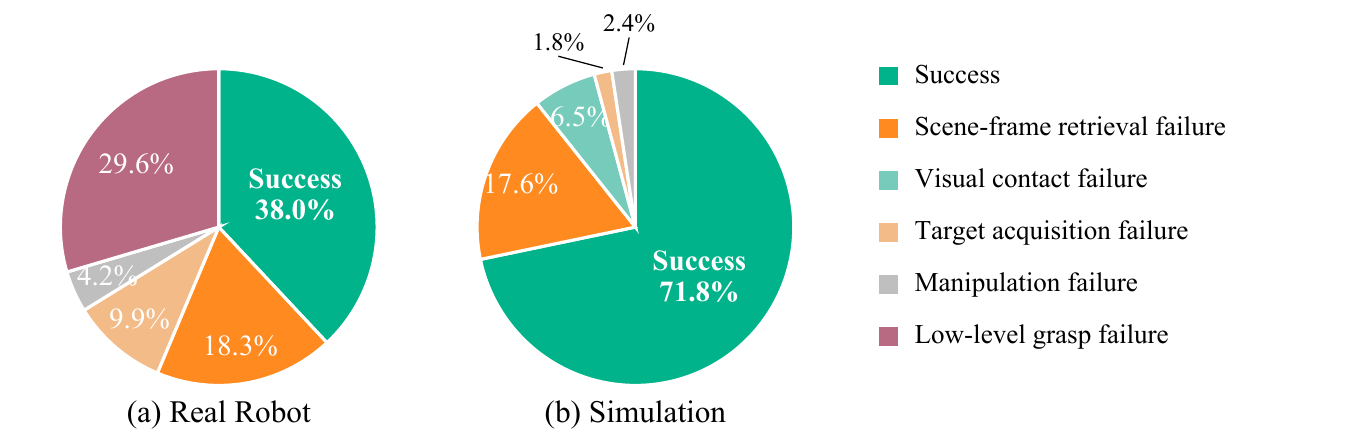}
  \caption{\textbf{Task outcomes.} Real-robot outcomes include low-level grasp failures, shown separately.}
  \label{fig:failure_modes}
\end{figure}

Low-level grasp execution accounts for the largest failure share on the real robot, while scene-frame retrieval is the largest failure source in simulation (Figure~\ref{fig:failure_modes}). Scene-frame retrieval and target acquisition also contribute to real-robot failures, highlighting the need for reliable manipulation skills and visual guidance during approach.

\newpage
\section{OWMM Training and Evaluation Hyperparameters}
\label{app:owmm_hyperparameters}

Table~\ref{tab:owmm_hyperparameters} summarizes the training and episode-evaluation configuration for the AIDA-OWMM policy used in the OWMM experiments.

\begin{table}[!ht]
  \centering
  \caption{\textbf{AIDA-OWMM model, training, and evaluation configurations.}
  The shared prompt and output format is defined in Section~\ref{sec:method:harness}; evaluation uses the trained LoRA adapter.}
  \label{tab:owmm_hyperparameters}
  \scriptsize
  \setlength{\tabcolsep}{5pt}
  \renewcommand{\arraystretch}{0.94}
  \resizebox{\textwidth}{!}{%
  \begin{tabular}{@{}p{0.30\linewidth}p{0.62\linewidth}@{}}
    \toprule
    Parameter & Value \\
    \midrule
    \multicolumn{2}{@{}l}{\emph{Base model and adapter}} \\
    Base VLM & Qwen3-VL-8B-Instruct \\
    Fine-tuning method & Supervised fine-tuning with LoRA \\
    LoRA rank / alpha / dropout & 64 / 128 / 0.05 \\
    LoRA target modules & \texttt{all} attention and feed-forward linear layers \\
    Rank-stabilized LoRA & Enabled \\
    Trust remote code & Enabled \\
    \midrule
    \multicolumn{2}{@{}l}{\emph{Training data and prompt schema}} \\
    Prompt template & \texttt{qwen3\_vl\_nothink} \\
    Stage inputs & Search: system prompt, scene key frames, current RGB, task, skill definitions, episode history, output schema. Navigation, pick, and place: system prompt, current RGB, task, skill definitions, episode history, output schema. \\
    Privileged inputs excluded from training and inference & Object mask, target/goal coordinates, projected navigation overlay, and privileged distance. \\
    Output fields & \texttt{state\_estimation}, \texttt{reasoning}, \texttt{action}, \texttt{action\_information}, \texttt{summarization} \\
    Cutoff length & 4096 tokens \\
    Image / video max pixels & 262144 / 16384 \\
    Preprocessing / dataloader workers & 16 / 4 \\
    \midrule
    \multicolumn{2}{@{}l}{\emph{Training optimization}} \\
    Per-device train batch size & 3 \\
    Gradient accumulation steps & 2 \\
    Effective batch size & $6 \times$ number of training GPUs \\
    Learning rate & $5.0 \times 10^{-5}$ \\
    Training epochs & 3.0 \\
    LR scheduler / warmup & Cosine / 0.03 warmup ratio \\
    Precision & bfloat16 \\
    Gradient checkpointing & Enabled \\
    Logging / checkpoint interval & 10 / 500 steps \\
    Checkpoint retention & 3 checkpoints \\
    Validation during training & Disabled \\
    \midrule
    \multicolumn{2}{@{}l}{\emph{Episode evaluation}} \\
    VLM decoding & 512 max output tokens, temperature 0.0, top-$p$ 0.5 \\
    Max VLM decisions per episode & 10 \\
    Episode history cap & 30 previous steps; no previous RGB history images \\
    Parallel evaluation & 20 workers across four GPUs \\
    Resume unfinished episodes & Enabled \\
    Pick timing at deployment & Selected by the VLM from its current observation \\
    Grasp threshold override & Habitat default \\
    \bottomrule
  \end{tabular}%
  }
\end{table}

\end{document}